\documentclass[a4paper]{article}
\usepackage[margin=25mm]{geometry}
\usepackage{amssymb}

\usepackage{amsmath}
\usepackage{url}
\usepackage{subfigure}
\usepackage{graphicx, xcolor}
\usepackage{amsfonts}
\usepackage[normalem]{ulem}
\usepackage{cite}

\usepackage{fancyhdr}

\usepackage{cite}

\usepackage{color}
\definecolor{darkgreen}{rgb}{0,0.5,0}

\newcommand{\sai}[2][]{%
   {\color{red}\sout{#2}}%
   \ifx&#1&%
      {}
    \else%
      \footnote{{\color{red}(sai):} #1}
    \fi%
}

\begin{document}
%
\title{Efficient Information Diffusion in Time-Varying Graphs through Deep Reinforcement Learning}

\author{Matheus R. F. Mendon\c{c}a,
        Andr\'{e} M. S. Barreto, and
        Artur Ziviani
\thanks{M. R. F. Mendon\c{c}a, A. M. S. Barreto and A. Ziviani are with the National Laboratory for Scientific Computing (LNCC), Petr\'{o}polis, RJ, Brazil. E-mails: \{mrfm,amsb,ziviani\}@lncc.br}
\thanks{M. R. F. Mendon\c{c}a is currently with DASA, S\~{a}o Paulo, SP, Brazil and A. M. S. Barreto is currently with DeepMind, London, UK.}}

\date{}



\maketitle

\begin{abstract}
Network seeding for efficient information diffusion over time-varying graphs~(TVGs) is a challenging task with many real-world applications. There are several ways to model this spatio-temporal influence maximization problem, but the ultimate goal is to determine the best moment for a node to start the diffusion process. In this context, we propose Spatio-Temporal Influence Maximization~(STIM), a model trained with Reinforcement Learning and Graph Embedding over a set of artificial TVGs that is capable of learning the temporal behavior and connectivity pattern of each node, allowing it to predict the best moment to start a diffusion through the TVG. We also develop a special set of artificial TVGs used for training that simulate a stochastic diffusion process in TVGs, showing that the STIM network can learn an efficient policy even over a non-deterministic environment. STIM is also evaluated with a real-world TVG, where it also manages to efficiently propagate information through the nodes. Finally, we also show that the STIM model has a time complexity of $O(|E|)$. STIM, therefore, presents a novel approach for efficient information diffusion in TVGs, being highly versatile, where one can change the goal of the model by simply changing the adopted reward function. 
\end{abstract}


\section{Introduction}

Network  seeding for efficient information diffusion in dynamic networks is an important and timely research topic that configures a challenging task with many real-world applications. This spatio-temporal influence maximization problem can be applied to different scenarios, ranging from social networks \cite{inf_node_1,fb_case,prop_behavior} to epidemiological analysis~\cite{epidemic}. It is a phenomenon that can be studied over different perspectives: (i)~one can be interested in halting a diffusion process, as it happens in an epidemic spreading, for instance; or (ii)~accelerating the diffusion to reach as many nodes as possible.
In both cases, it is crucial to correctly identify the most efficient spreader (seeding) nodes, which are nodes with high diffusion capabilities in a given network at a given time frame.

Identifying efficient spreader nodes is no easy task due to the fact that real-world networks are usually dynamic, represented by time-varying graphs~(TVG), so the connectivity of a node may constantly change. This makes it harder for identifying an important node to start an information diffusion over a series of time steps. As previously stated~\cite{time_cent}, in a TVG, it is difficult to identify if a node is currently in its apex of importance or if this importance will still grow over the following time steps. At the same time, if we wait too much for a node to reach its peak of importance, we might miss its actual importance peak. Note that another issue is how to define its importance from a time perspective: It depends on the problem at hand, and many approaches proposed different centrality measures for diffusion processes in TVGs~\cite{time_evol,kshell,ranking_key,time_cent}.

In this work, we introduce a novel approach for the spatio-temporal influence maximization problem, i.e. the problem of finding a node and a time instant for the seeding activation that leads to an efficient information diffusion in a TVG. 
In this sense, we propose the Spatio-Temporal Influence Maximization (STIM), a model trained with Reinforcement Learning~(RL) and Graph Embedding~(GE) that learns to identify a suitable moment to release a given information from nodes such that its propagation throughout the TVG is maximaized. It also identifies the best node to start a diffusive process at each time step. The STIM is trained using a series of artificial TVGs, and can then generalize to real-world TVGs. We show through a series of experiments that the STIM network is capable of efficiently diffusing information over a TVG. We compare our approach with a greedy agent and a highly specialized (oracle) agent, where the former uses a simple and low-cost strategy, while the latter uses a strategy built using intrinsic information of the creation process of the artificial TVG. The STIM manages to outperform the greedy agent and achieves comparable results to the oracle agent, showing that it manages to learn complex patterns hidden in the artificial TVGs without any supervised data or handcrafted strategies~(like the oracle agent). Finally, we show that the STIM network has a time complexity of $O(|E|)$, allowing it to be executed in large graphs without requiring a prohibitive computational cost. The source code for the STIM model is publicly available on GitHub.\footnote{https://github.com/MatheusMRFM/STIM} 

The remainder of this paper is organized as follows. In Section~\ref{sec:basic}, we review the basic concepts of RL, GE, and TVGs. We then discuss related work in Section~\ref{sec:related}. In Section~\ref{sec:method}, we introduce the proposed STIM model. We then evaluate the proposal in Section~\ref{sec:result}. Finally, the conclusions and future work perspectives are presented in Section~\ref{sec:conclusion}.

\section{Background}
\label{sec:basic}

The STIM framework is built using Reinforcement Learning~(RL), Graph Embedding~(GE) techniques, and Time-Varying Graphs~(TVG). 
Therefore, in this section, we briefly revise basic concepts in RL, GE, and TVGs, important to following the paper. 

\subsection{Reinforcement Learning (RL)}
\label{sec:rl}

In RL, an agent learns to interact with the environment through experience. The learning process occurs over discrete time steps. At time step $t$, the agent: (i)~obtains an observation $o_t$ of the environment; (ii)~determines the current state $s_t$ based on $o_t$; (iii)~performs an action~$a_t$; and (iv)~the environment returns a reward signal $R_t$. This process repeats until a termination condition is satisfied. The main goal is to discover a mapping from actions into states, i.e., a policy~$\pi$, that maximizes the expected return $G_t = \sum_{i=0} \gamma^i R_{t+i+1}$, where $\gamma$ is called the discount factor.

\textit{Markov Decision Processes} (MDPs) are commonly used in RL. An MDP is defined as $(\mathcal{S}, \mathcal{A}, p, r, \gamma)$, where $\mathcal{S}$ is the set of possible states, $\mathcal{A}$ is the set of actions, $p$ is the state transition distribution function $p(s'|s,a)$ for $a \in \mathcal{A}$, $s'\in \mathcal{S}$ and $s\in \mathcal{S}$, while $r$ is the reward function $r(s,a,s')$ that determines the reward for performing action $a\in \mathcal{A}$ while in state $s\in \mathcal{S}$ and transitioning to state $s'\in \mathcal{S}$. 
To encourage the agent to explore the state space, RL methods usually adopt an $\epsilon$-greedy policy, where the agent selects a random action $\epsilon$\% of the time, while the remaining ($1-\epsilon$)\% of the actions are selected according to $argmax_a Q^{\pi}(s,a)$. Factor $\epsilon$ is called the exploration rate.

The traditional action-value function $Q(s,a)$ predicts the exact value of expected future rewards for a given state-action pair $(s,a)$. However, this setting is not ideal for non-deterministic environments, where the state transition distribution function $p(s'|s,a)$ is highly stochastic. Such environments are relatively common in real-world scenarios, thus a specific RL formulation for these conditions is ideal. The \textit{Categorical Algorithm}~\cite{dist_ql} presents a distributional view of RL, i.e., it predicts a distribution of possible values for $Q(s,a)$ instead of its exact value. This is useful because it allows RL to behave more steadily in stochastic environments, where the RL agent is bound to select optimal actions and not receive a high reward as expected. The \textit{categorical algorithm} first defines a set of $n_a$ \textit{atoms}, which represent a discretization of the possible returns. The i-th atom is given by $z_i = V_{min} + i \Delta z$, where $\Delta z = \frac{V_{max} - V_{min}}{n_a - 1}$ and $V_{max}$ and $V_{min}$ are the maximum and minimum expected sum of returns, respectively.

Given a set of atoms, the goal is a function $\rho(s,a)$ that maps state-action pairs into a probability distribution for the expected sum of rewards over the set of atoms $z_i$. Given that $\rho$ is parameterized by a function $\theta(s,a)$, 
we have that:

\begin{equation}
	\rho_i (s,a) = \frac{e^{\theta_i(s,a)}}{\sum_j e^{\theta_j(s,a)}} \forall i,j \in [0, n_a-1].
	\label{eq:pz}
\end{equation}

With $\rho(s,a)$, we can compute a new $Q(s,a)$, given by:

\begin{equation}
	Q(s,a) = \sum_i z_i \rho_i (s,a).
	\label{eq:dist_q}
\end{equation}

Since the goal is to find an optimal distribution function $\rho(s,a)$, we can compute the actual distribution obtained through experience, given by $m \in \rm I\!R^{n_a}$, and compare the difference between these two distributions in order to determine the loss function. The loss function used is the cross-entropy loss $L_{ce}$ between $\rho(s,a)$ and $m$. Please refer to~\cite{dist_ql} for more details regarding the Categorical Algorithm. 


\subsection{Graph Embedding (GE)}
\label{sec:ge}

Node and graph embedding represents a set of methods capable of encoding topological information and node characteristics into low dimension vectors. This allows each node to be represented by an $F$ dimensional vector that encodes a set of features and its connectivity information. 

In this section, we briefly revise two well-established GEs: 

\begin{itemize} 
	\item GCN~\cite{gcn} is a graph embedding method that builds latent representations of a node by using a set of node features and the underlying graph structure as input; 
	\item In Structure2Vec~\cite{s2vec}, each node is embedded using a set of node's features and the embedding of all of its neighborhood. This is also an iterative method that computes a new representation of the embedding matrix $H$ over a set of $L$ layers. The node embedding $H_i^l$ of node $i$ in layer $l$ is built using the topological information of its $l$-hop neighborhood. By using $L = d_m$, where $d_m$ is the diameter of the graph, the resulting embedding matrix $H^L$ considers the connectivity properties of the entire graph. Note that $d_m$ is usually small for real-world networks, so the number of layers used by this method is also small. The embedding $h_i^{l+1}$ of a node $i$ at layer $l+1$ is given by:

	\begin{equation}
		h_i^{l+1} = \sigma \left(x_i, \sum_{n\in \mathcal{N}(i)} h_n^l \right),
		\label{eq:embed_simple}
	\end{equation}
	
	\noindent where $\mathcal{N}(i)$ is the neighborhood of node $i$, $x_i$ is the set of features of node $i$ (given by row $i$ of the feature matrix $F_M$), and $\sigma$ is any nonlinear function. Figure~\ref{fig:architecture}(a) shows the full architecture of the Structure2Vec model, where we better detail it in Section~\ref{sec:method}.
\end{itemize}

\subsection{Time-Varying Graphs (TVG)}
\label{sec:tvg}

There are different ways to model a TVG~\cite{tvg,MAG2}. Here, we consider a TVG as a set of snapshots of the TVG over discrete time instants. Therefore, we model a TVG $\mathbb{G}$ as a set of static graphs $G_i=(V_i,E_i)$ that represents the TVG at time instant $i$, where $V_i$ is the set of nodes and $E_i$ is the set of edges at time step $i$. We also consider in this work that a TVG maintains the same set of nodes throughout different time instants, that is, $V_i = V_j \forall i,j \in [1,T]$, $T$ being the final time step of $\mathbb{G}$. The number of nodes in a graph is represented here by $N_i = |V_i|$. Since $V_i = V_j \forall i,j \in [1,T]$, we consider a single $N = |V_i| \forall i \in [1,T]$.

\section{Related Work}
\label{sec:related}

In this section, we describe some of the research focused on understanding and controlling a diffusion process in a network and how to use RL for graph-based problems. There are three main approaches adopted for controlling the information diffusion: (i)~build local centrality measures that only consider the vicinity of a node and its current time step in order to identify its importance in a diffusive process; (ii)~time-based centrality measures, which considers the past connectivity of a node; and (iii)~influence maximization approaches. We also present how RL and GE have been previously used together.

\subsection{Local Centrality Metric}

An important aspect in diffusion processes over a TVG is how to select important nodes to propagate an information in a way this information spreads efficiently in the network. Kim and Yoneki~\cite{inf_node_1} addresses this problem by elaborating a set of node selection strategies in order to maximize the reach of a given information. The authors present an adaptation to the \textit{Independent Cascade} (IC) model~\cite{ind_cascade} for information diffusion in networks: Given a limited time frame and a set of nodes that starts with a given information, the goal is to select the neighboring nodes that maximize the reach of this information. These strategies range from random selection to more complex mechanisms, such as choosing the neighbors with the highest volumetric centrality~\cite{vol_cent}. 
Berahmand et al.~\cite{new_cent} define a new node centrality measure capable of identifying good nodes to spread an information based on first and second-order information. Their new centrality measure uses the clustering coefficient of a node and the sum of clustering coefficients of its neighborhood, along with its degree, in order to decide if a node is a spreader of information or not. 
For other local centrality metrics proposed in the literature to identify important spreader nodes, please check~\cite{local_3, local_4}.

\subsection{Time-Based Node Centrality}

A different way to approach the influence maximization problem is to build a time-based node centrality that measures the importance of a node in the TVG according to a specific characteristic. For an influence maximization, one can define a time-based centrality that measures the diffusion capabilities of a node considering its connectivity behavior in the TVG. 

Wang et al.~\cite{ranking_key} present the Temporal Degree Deviation centrality measure. This centrality basically computes the degree deviation of a node along with several time steps. A high degree deviation indicates that a node is an important spreader in the network. 

Costa et al.~\cite{time_cent} present two different time centrality metrics: \textit{cover time} (CT) and \textit{time-constrained coverage} (TCC). The CT centrality $ct(i,t,\lambda)$ of node $i$ indicates the number of time steps required for a diffusive process starting from node $i$ at time step $t$ reaches a specific fraction $\lambda$ of nodes in the graph. Hence, lower values of $ct(i,t)$ indicates a more central node at time step $t$. 
On the other hand, the TCC centrality $tcc(i,t,\lambda,\sigma)$ indicates the fraction $\lambda$ of nodes reached after starting a diffusive process from node $i$ at time step $t$ for a fixed number of steps $\sigma$. Therefore, high values of $tcc(i,t,\lambda,\sigma)$ indicates that a node is more central. 

Trust-based Most Influential node Discovery~(TMID)~\cite{node_group} 
presents the concept of trust connections between nodes and the influence of a node over another. These attributes are determined by analyzing how other nodes react to a diffusion process started by a specific node~$i$.
Based on these reactions, TMID computes the influence of $i$ over other nodes, as well as how trustworthy~$i$ is in the network. TMID finally elects a set of top-$k$ influential nodes to propagate an information in the network.

\subsection{Influence Maximization Approaches}

The Influence Maximization (IM) problem, first proposed by Kempe et al.~\cite{max_spread}, studies how to best spread an information in a static network over several discrete time steps. For this purpose, a set of $k$ nodes (seeds) are activated, and once active, these nodes start spreading the information. The goal is to determine the best $k$ nodes to be activated such that it maximizes the number of influenced nodes. Several variants of the IM problem have been reported in the literature, such as IM with limited budget for activating nodes~\cite{node_sur,Dyn_IC}, to time-constrained IM~\cite{time_im}. 
Such IM approaches only deal with static graphs.

\subsection{Graph Embedding and Reinforcement Learning}

The work presented by Khalil et al.~\cite{combinatorial} is the only approach found that mixes graph embedding and reinforcement learning, although not applied to the problem of diffusion in TVGs. The authors present a novel approach for merging graph embedding and reinforcement learning in order to solve combinatorial graph problems, such as the traveling salesman problem. 

\subsection{Contributions}

To the best of our knowledge, our approach is the first method that tackles a spatio-temporal IM for TVGs. Former methods only considered the IM problem for static graphs, which does not efficiently represent many real-world dynamic networks. Although here we disregard a limited budget IM problem, we show in Section~\ref{sec:conclusion} how to model this problem using our proposed approach.


\section{Spatio-Temporal Influence Maximization}
\label{sec:method}

\begin{figure*}[htp]
	\centering
	\subfigure[Structure2Vec Node Embedding]{\includegraphics[scale=0.9]{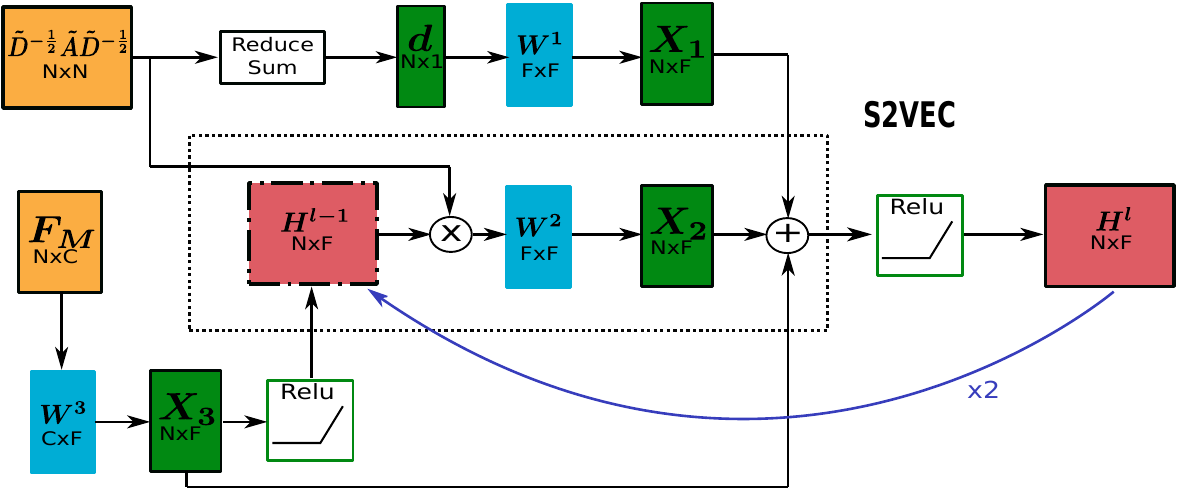}
		\label{subfig:s2v}}
	\subfigure[STIM Framework]{\includegraphics[scale=0.9]{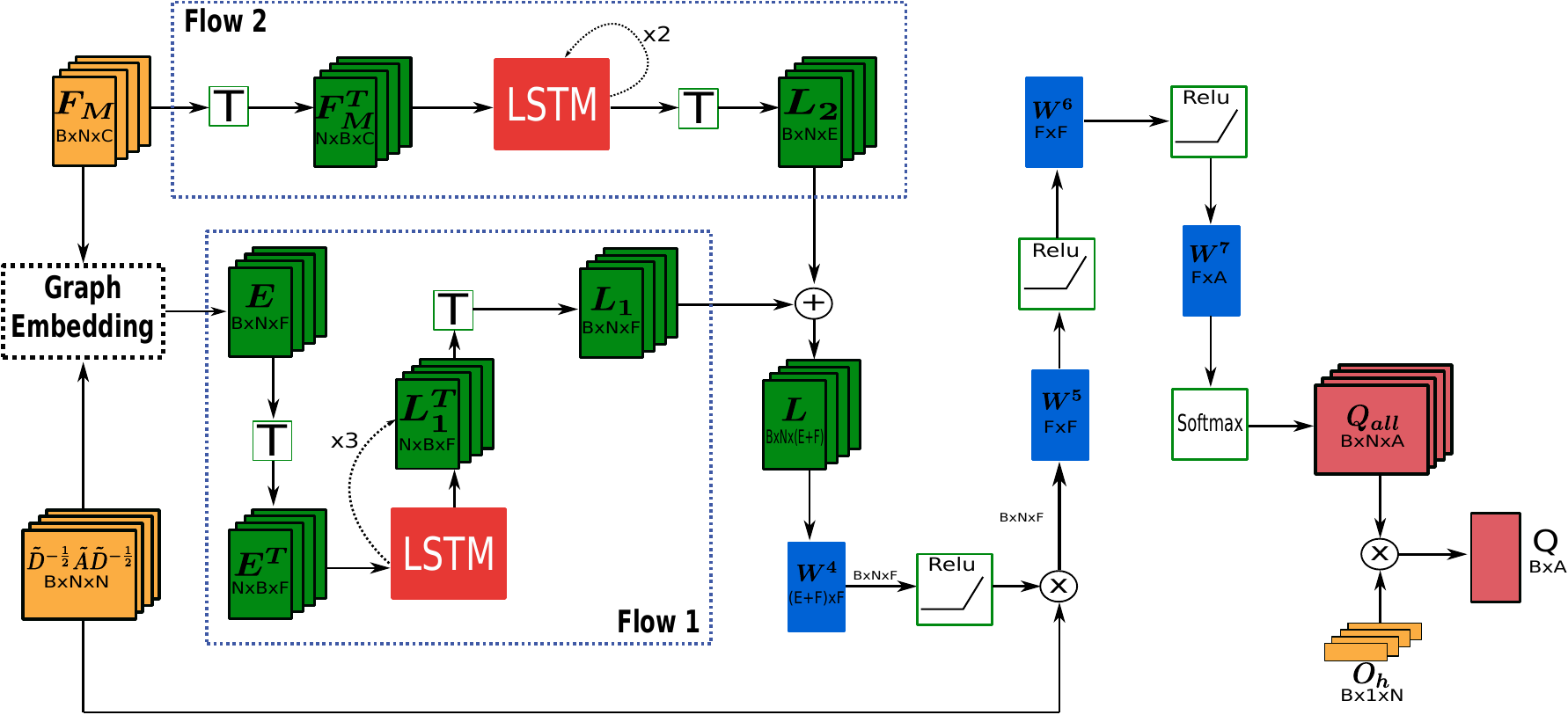}
		\label{subfig:STIM}}
	\caption{Architecture of the used model: (a)~Structure2Vec variation used for the graph embedding; and (b)~the proposed Spatio-Temporal Influence Maximization framework~(STIM). The inputs are represented in yellow, intermediate matrices are in green, hidden layers are in blue, and output matrices are in light red. Each batch contains $B$ contiguous snapshots of the same TVG. That is, for each batch, we pass a set of $B$ feature matrices $F_M$ and normalized adjacency matrices $\Tilde{D}^{-\frac{1}{2}} \Tilde{A} \Tilde{D}^{-\frac{1}{2}} H^{(l)} W^{(l)}$. $N$	is the number of nodes in the TVG being processed in the batch, $C$ and $F$ are the number of features used in the feature matrix and embedding matrix, respectively, and $Q_{all}$ is the distributional action-value function $Q(s,a)\in \rm I\!R^{1xA}$  for all $N$ nodes of the TVG. To output only the $Q(s,a)$ value for the action being processed, we also pass a set of $B$ one-hot vectors $O_h$ indicating the node used as action at each of the $B$ time steps. The final output of STIM is the $Q$ function for states $s^t, s^{t+1}, \ldots, s^{t+B-1}$ after selecting nodes $O_{h}^t, O_{h}^{t+1}, \ldots, O_{h}^{t+B-1}$ at the respective states.}
	\label{fig:architecture}
\end{figure*}

Here we present the \textit{Spatio-Temporal Influence Maximization} (\textit{STIM}), an RL model capable of selecting a suitable moment to start a diffusion process from a given node. We argue that the STIM can also be tuned for other diffusion control problems in TVGs by just changing the reward function of the model. Therefore, the STIM can also be considered a more general framework for diffusion problems in TVGs. The STIM network is also a novel approach to the problem of diffusion in TVGs: As shown in Section~\ref{sec:related}, to the best of our knowledge, there is no method in the literature that uses a machine learning approach to learn the best moment to start an efficient diffusion process in a dynamic network. Thus, the STIM can be considered a new branch in the possible approaches for this problem, focused on reinforcement learning and graph embedding. However, it is important to note that the STIM network is inspired by some existing approaches: the combination of RL and GE presented in \cite{combinatorial}, and the problem definition for diffusion in TVGs presented in~\cite{time_cent,evol_dyn}.

In this section, we first present how we model the problem of diffusion in TVGs, describe the details of the STIM net, and also show how the model is trained and tested.

\subsection{Diffusion in TVGs}
\label{subsec:dif}

The goal of the STIM framework is to identify and select the best nodes to start a diffusion process at a given time step~$t$ in order to maximize the reach of an information in the dynamic network. As defined in Section~\ref{sec:tvg}, we consider that the number of nodes is kept fixed across different time steps of the same TVG, although this number of nodes may vary between different TVGs, as the STIM works for TVGs with an arbitrary number of nodes. Further, as the connections in the TVGs may be constantly changing, so the degree $d_i^t$ of node $i$ at time step $t$ may be different from $d_i^{t-1}$ or $d_i^{t+1}$. 

Here, we consider the scenario where an information (data) must be diffused in a TVG. The objective is to diffuse this information to the maximum possible number of nodes. However, the information can only be propagated when a node that retains the information is activated, sending the information to all its immediate neighbors. Only one node can be activated at each time step, and the node can only be activated if it currently has the information in its power. 
An external agent, referred from here on simply as \textit{agent}, is responsible for selecting which node will be activated at each time step. We model the diffusion process in a TVG according to the following rules:

\begin{itemize}
	\item Each node has three possible states:
	\begin{enumerate}
		\item \textbf{Neutral:} when the node does not have any contact with the data being propagated;
		\item \textbf{Retain:} when the node received the information and is now retaining it. In this state, the node can be activated by the agent controlling the diffusion process (STIM, for example). Also, while in this state, the information is not propagated to its neighbors; 
		\item \textbf{Informed:} when a node is activated, it propagates the information for one time step and then drops the information. At this point, the node is considered to be informed, although it can not propagate the data any longer.
	\end{enumerate}
	The only two possible node state transitions are: \textit{neutral}$\rightarrow$\textit{retain} and \textit{retain}$\rightarrow$\textit{informed}.
	
	\item When a node propagates the information, its neighbors have a probability $\phi$ of accepting the information. $\phi$ is a parameter of the diffusion model and can be defined as a constant or as a function that considers certain features of each node. We better detail this parameter in~Section~\ref{sec:test_net};
	
	\item At the beginning of the diffusion process, a set of nodes $P$ are in the \textit{retain} state, while the remaining nodes are in the \textit{neutral} state. There are no nodes in the \textit{informed} state at the start of the process;
	
	\item The nodes can only start being activated after $k$ steps (counting from time step 1). After $k$ steps, the agent has $q$ time steps to activate the nodes. Given these $q$ time steps, the diffusion process ends and the number of \textit{informed} nodes gives the score of the agent for that particular simulation. The values of $k$ and $q$ are not fixed and are defined randomly at the start of each simulation, i.e., for two different simulations using the same TVG $\mathbb{G}$, the values of $k$ and $q$ may differ;
	
	\item The goal is to maximize the number of \textit{informed} nodes at the end of each simulation.
	
\end{itemize}

The diffusion process considered here considers an information that must be propagated through the network. Nevertheless, note that this problem can also be thought of as an epidemiological simulation, where we want to identify the nodes with the greatest spreading capacity in the network. 

The agent controlling the activation of the nodes can be defined in several ways. Here, we propose the STIM agent, a reinforcement learning approach that learns a sub-optimal policy $\pi$ capable of selecting the node with the greatest spreading capacity, in order to disseminate the information more efficiently in the TVG. In the remainder of this section, we detail how the STIM accomplishes this task.

\subsection{Graph Embedding in the STIM}
\label{subsec:s2vec}

The first step of the STIM agent is to embed all nodes of the graph, as we can seen in Figure~\ref{subfig:STIM}. The STIM is not restricted to a specific graph embedding method. The only requirement is that the output of the selected graph embedding method should be a node embedding matrix $H \in \rm I\!R^{NxF}$ for each graph $G_t$ of the TVG $\mathbb{G}$, where $N$ defines the number of nodes in $\mathbb{G}$ and $F$ is the size of the embedding. For this work, we used the Structure2Vec model with some minor tweaks as the graph embedding method for the STIM. We also tested it with GCN, but the Structure2Vec presented better results. We now describe some of the implementation details for our version of the Structure2Vec. Figure~\ref{subfig:s2v} presents a schematic of the used GE model.

The formal definition of the Structure2Vec presented in Eq.~\ref{eq:embed_simple} only defines how to embed a single node. In order to make the implementation more optimized, we formulate the Structure2Vec using matrix operations, allowing us to compute the entire embedding matrix $H^l$ at layer $l$ in a single set of operations (Figure~\ref{subfig:s2v}). Following the approach of Hanjun et al.~\cite{combinatorial}, we used an extra weight matrix to consider the connection weights of the graph, depicted by the upper flow in Figure~\ref{subfig:s2v}. We also used the normalized adjacency matrix with self-loops used in the GCN method $\Tilde{D}^{-\frac{1}{2}} \Tilde{A} \Tilde{D}^{-\frac{1}{2}}$~(check~\cite{gcn} for more details) instead of the conventional adjacency matrix $A$, given that the former matrix performed better within our preliminary results. The parameterization of the model is then given by:

\begin{eqnarray}
	d_i &=& \sum_{j=1}^{N} (\Tilde{D}^{-\frac{1}{2}} \Tilde{A} \Tilde{D}^{-\frac{1}{2}})_{ij} \\
	X_1 &=&  d W^1 \\
	X_3 &=& F_M W^3 \\
	H^0 &=& relu(X_1 + X_3) \\
	H^{(l+1)} &=& relu \left( A H^l + X_1 + X_3 \right),
	\label{eq:s2vec_parameter}
\end{eqnarray}

\noindent where $H^l$ is the embedding matrix at layer $l$, $d \in \rm I\!R^{Nx1}$ is the degree vector, and $W^1, W^2 \in \rm I\!R^{FxF}$ and $W^3 \in \rm I\!R^{CxF}$ are the weight matrices. We used 2 layers for the graph embedding, therefore $H^2$ represents the final embedding matrix.

The graph embedding method uses the feature matrix $F_M \in \rm I\!R^{NxC}$ as one of its inputs, which gives the value of $C$ raw features of each node. These features must be defined beforehand, and here we used four features ($C=4$): (i)~the normalized degree of the node; (ii)~its eigenvector centrality; (iii)~the mean; and (iv)~standard deviation of the degree of a node over the last 5 time steps. All four features have a low computational cost. The most costly feature is the eigenvector centrality of a node, but during our experiments, we noticed that this feature is not crucial to the success of the method. Therefore, in the case computational time is important, one can use only the remaining three features.

Note that other features may be also used by the STIM network, that is, these four features are not fixed. The features used may also use certain task-specific features for each node. For example, if considering a disease propagating in a network where each node represents a person, then we can use the age of each person as a feature. Therefore, these $C$ features are actually part of modeling problem and may vary depending on the problem at hand.

\subsection{STIM}

As previously stated, the STIM network is an RL model capable of selecting nodes to start a diffusion process in a TVG in order to maximize the information propagation. The layout of STIM is presented in Figure~\ref{subfig:STIM}. The input to the model is a set of graphs that represent snapshots of a TVG over sequential time steps. The number of these snapshots passed to the model depends on the used batch size. Therefore, for a batch size $B$, we pass the graphs $G_t, G_{t+1}, \ldots, G_{t+B-1} \in \mathbb{G}$, where $t$ is the time step of the first graph in the batch.  

The RL approach used by STIM is a categorical algorithm 
We used this approach over the conventional Q-Learning, for example, because we are dealing with a highly stochastic simulation, where sometimes the number of $informed$ nodes may increase very little even after selecting an optimal node, which can mislead the agent during its learning process. Therefore, learning the distribution of future rewards instead of the exact values of $Q(s,a)$ proved to be an important step.

The first step of the STIM model is to embed all nodes of all graphs into the batch. Each graph $G_i \forall i \in [t,t+1,\ldots,t+B]$ is processed through a graph embedding method, which outputs a series of $B$ graph embedding matrices $H_i \in \rm I\!R^{NxF}$. As previously stated, STIM works with any graph embedding method that outputs an embedding matrix. For this work, we used a modified version of the Structure2Vec method (see Section~\ref{subsec:s2vec}). The result of this step is an embedding matrix $H \in \rm I\!R^{BxNxF}$, containing the embedding matrix $H_i$ of each snapshot in the batch.

In a TVG, the pattern of how the connections of a node change may indicate several trends and also helps to predict its future behavior. Therefore, capturing these patterns is crucial in order to determine nodes with high propagation capabilities. To this end, we adopt recurrent layers that are able to capture long and short term patterns (LSTM layer). In short, an LSTM layer receives a batch of time series in the shape $b \times l \times m$, where $b$ is the batch size, $l$ the length of each time series, and $m$ is the representation of an instance of a time series. The LSTM then computes a new state representation for the next instance of the time series. For example, in weather forecast, we can pass a single time series ($b=1$) to an LSTM comprised of the temperature and humidity ($m=2$) of the previous 10 days ($l=10$). The LSTM then returns the prediction of both features for the following day, that is, for each instance $i$ in the time series, it predicts temperature and humidity for the instance $i+1$. Following this idea, we want to pass the past embeddings of each node to the LSTM layer, in order to predict the embedding of each node in the following time step. Since each batch contains $B$ contiguous snapshots, then our time series $l=B$. Given that the embedding of each node is independent to other nodes (their embeddings already account for their connectivity and their neighbors), we consider the time series of the embeddings of each node as a separate instance in the batch, that is, the batch size passed to the LSTM is $b=N$. Each instance of the time series is the node embedding computed by the graph embedding method, that is, $m=F$. Hence, we must pass a matrix with dimensions $N \times B \times F$. To do this, we simply compute the transpose of the embedding matrix $H$ and feed it to the LSTM layer. The output of this layer is a matrix with dimensions $N \times B \times F$, which represents the prediction of the embedding of each node in the next time step based on the behavior of each node along with the previous $B$ time steps. As a final step, we reverse the transpose operation, transforming the dimensions of the output of the LSTM from $N \times B \times F$ back to $B \times N \times F$ and storing the transformation result in matrix $L_1$.

As important as the node embedding pattern is its pattern of raw features~$C$, i.e., how the basic features (from the feature matrix $F_M$) change over time for each node. Thus, we use a second processing flow, where we want to encode the changing pattern of raw features in each node. We use the same resource as in the original flow, transpose the feature matrix $F_M \in \rm I\!R^{BxNxC}$ to $F_{M}^T \in \rm I\!R^{NxBxC}$, and feed this matrix to a different LSTM layer. We then create a high order representation of the pattern of each node using $E$ features, i.e., the output of this second LSTM layer is given by $L_{2}^T \in \rm I\!R^{NxBxE}$, which is then transposed back to  $L_{2} \in \rm I\!R^{BxNxE}$. We then concatenate both matrices $L_1$ and $L_2$, given by flow 1 and flow 2, respectively, resulting in matrix $L \in \rm I\!R^{BxNx(F+E)}$. 

After computing matrix $L$, we perform a second graph embedding, but using a simpler approach. First, we multiply each matrix $L_i \in \rm I\!R^{Nx(F+E)} \forall i = [1,\ldots,B]$ by a weight matrix $W^4 \in \rm I\!R^{(F+E)xF}$, followed by the Relu activation function. The resulting matrix with dimensions $B \times N \times F$ is considered the new feature matrix (with $F$ features for each node over each snapshot). We then multiply each of the $B$ feature matrices with the $B$ normalized adjacency matrices $\Tilde{D}^{-\frac{1}{2}} \Tilde{A} \Tilde{D}^{-\frac{1}{2}}$. This resulting matrix is then multiplied by the weight matrices $W^5, W^6 \in \rm I\!R^{FxF}$ and $W^7 \in \rm I\!R^{FxA}$, where $A = n_a$ is the number of atoms used by the categorical algorithm. 
A sotfmax is applied over the resulting matrix, with dimensions $B \times N \times A$. The result is the matrix $Q_{all} \in \rm I\!R^{BxNxA}$, which stores the distribution over the possible sum of future rewards for each of the $N$ possible actions, where each action is selecting a node to start a diffusion process. 

\subsection{Reward Function}

The reward function is crucial in any RL application. This function determines the behavior of the agent, given that its goal is to maximize the rewards received. A trivial reward function is to reward the agent based on the percentage of new nodes $informed$ after executing an action. For example, if at time step $t$ the TVG had 5\% of $informed$ nodes and then at time step $t+1$ there were 9\% of $informed$ nodes after selecting node $i$ to diffuse its information, then the agent receives a reward of $9-5\% = 4\% = 0.04$. Although simple, this reward managed to obtain relatively good results. 

We tested several other different reward functions, but ultimately, the best one (which was used to generate the results reported in this work) is what we call the \textbf{influence reward}. The aim of the influence reward is to reward actions that increase the \textit{influence} count of the TVG, which is the number of $neutral$ nodes that are neighbors to at least one node in the $retain$ state. Therefore, the influence count indicates the number of possible nodes that are susceptible to receiving the information~(not all at the same time step though).  This is important because it helps the agent to understand that sometimes it is more advantageous to select a sub-optimal node in case the optimal one is currently in a rising degree pattern. For example, if a node $i$ in the $retain$ state is currently increasing its degree for the past three time steps and the STIM finds that this node may continue to grow its degree, then it is better to leave this node in the $retain$ state and select it only wdhen it reaches its maximum degree, even if node $i$ is the best candidate at a given time step.  The influence reward is comprised of two terms:

\begin{itemize}
	\item \textbf{Diffusion:} represents the difference of percentage of $informed$ nodes between two time steps, as previously detailed;
	\item \textbf{Influence Count:} represents the difference of percentage of the influence count between two time steps.
\end{itemize}

Both terms of the influence reward are based on the difference between certain percentages. These values may vary depending on the size of the graph. To avoid this variance, we normalize both terms according to:

\begin{equation}
	j_{norm} = w_j * \frac{j - min_j}{max_j - min_j},
	\label{eq:norm_term}
\end{equation}

\noindent where $j$ represents the value of the diffusion or influence count terms, $w_j$ is the weight we give to each term, and $min_j$ and $max_j$ are the minimum and maximum values of $j$ observed over the duration of the training. With this normalization, the values of each term tend to be more expressive (closer to 1). For our experiments, we found that the best set of weights for each term were $w_{dif} = 1.0$ for the diffusion term and $w_{inf} = 0.5$ for the influence term. The final reward is then given by the sum of both normalized and weighted terms. Finally, to guarantee that the final reward is within the range [0,1], we apply a final normalization, given by:

\begin{equation}
	r_{norm} = \frac{dif_{norm} + inf_{norm}}{w_{dif} + w_{inf}},
	\label{eq:norm_r}
\end{equation}

\noindent where $dif_{norm}$ and $inf_{norm}$ are the normalized values (Eq.~\ref{eq:norm_term}) for the diffusion and influence count terms.

\subsection{Artificial TVGs}
\label{sec:artificial}

\begin{figure}[htp]
	\centering
	\subfigure[TVG from the training set]{\includegraphics[scale=0.3]{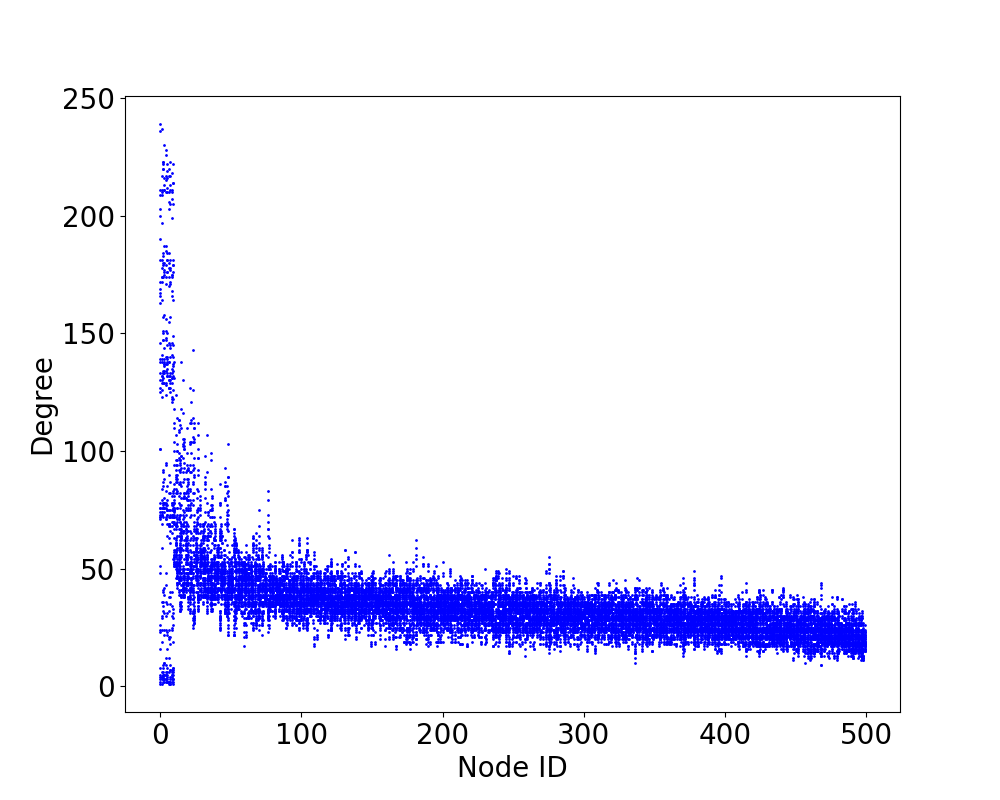}
		\label{subfig:train}}
	\subfigure[E-mail real-world TVG]{\includegraphics[scale=0.3]{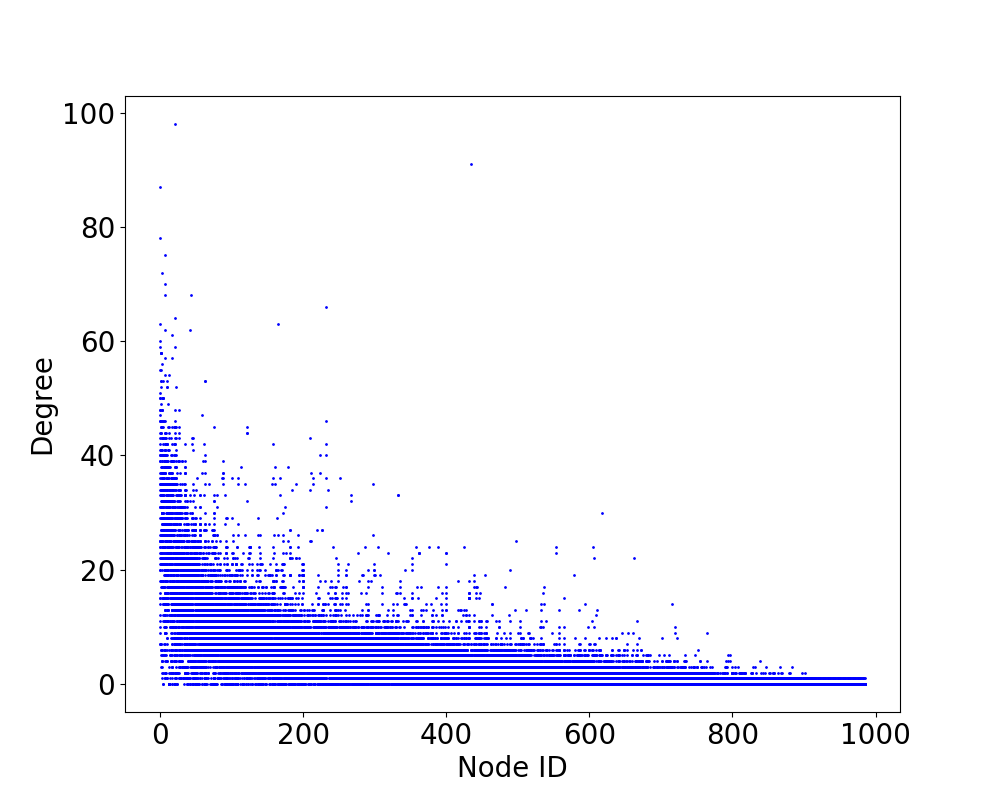}
		\label{subfig:email}}
	\caption{Scatter plot showing how the degree centrality of each node fluctuates in a TVG. The X-axis represents the id of each node (for example, the graph shown in (a) has approximately 500 nodes). The Y-axis shows the degree of each node, where each point represents the degree of a given node in a different time step.}
	\label{fig:scatter}
\end{figure}

The STIM network is capable of analyzing the connectivity behavior of a node in order to decide if it is a spreader node or not. In order to test this capability, we developed a set of artificial TVGs capable of testing this feature. These artificial graphs are used to train the STIM network. 

The main feature of these artificial TVGs is that a certain percentage of nodes have a cyclic behavior, where these nodes start with low connectivity (low degree) and then gain new connections at each new time step until they reach a high connectivity value. After that, these nodes start losing connectivity, until they reach a low degree, restarting the cycle. We call these nodes \textit{cyclic} nodes. If a cyclic node  is increasing its degree for each time step, we say that it is in a \textit{growth stage}, while a \textit{shrinking stage} is when the \textit{cyclic} node is losing connections.

The artificial TVGs have the following characteristics:

\begin{itemize}
	\item Each TVG has between 500 to 1000 nodes and between 30 to 60 time steps;
	
	\item The graph $G_1 \in \mathbb{G}$ representing the first snapshot of the TVG is created using the scale-free network model~\cite{barabasi}, which means that there are hubs in the network;
	
	\item A small percentage (2\%) of the nodes in the TVG are assigned to be \textit{cyclic};
	
	\item Between each time step, we add small perturbations over the connections of normal nodes and greater perturbations over \textit{cyclic} nodes. The perturbations over a normal node $i$ are performed by simply adding or deleting certain edges that have $i$ in one of its ends. The probability of these perturbations is chosen randomly for each TVG created, with a value between 1\%-7\%. For the \textit{cyclic} nodes, these perturbations depend on the current stage of the node: If it is in a \textit{growth stage} in a given time step, then it receives $p_c$\% of new connections, while losing $p_c$\% connections if its in the \textit{shrinking stage}. Here, we used $p_c$ = 15\%;
	
	\item Given the nature of the artificial graphs created, we can categorize each node into 5 classes:
	\begin{itemize}
		\item \textbf{0:} normal nodes with low connectivity (non-hub with a normalized degree below 0.4) with no connection to nodes of type 1;
		\item \textbf{1:} \textit{cyclic} node with low connectivity (normalized degree below 0.4). When its connectivity reaches 0.4 or higher, it becomes a node of type 4;
		\item \textbf{2:} similar to nodes of type 0, with the difference being that these nodes are connected to at least one type 1 node;
		\item \textbf{3:} normal nodes with high connectivity (normalized degree $\geq$ 0.4). These are the non-cyclic hubs;
		\item \textbf{4:} \textit{cyclic} node with high connectivity (normalized degree $\geq$ 0.4). When these nodes lose connectivity and reach a normalized degree below 0.4, they become a type 1 node;
	\end{itemize}
	
	\item The number of valid time steps where the agent can select nodes to start a propagation is given by $p_v$\% of the total number of time steps in the TVG. The starting time step is given by $t = \frac{T}{2}$, where $T$ is the number of time steps in the given TVG. Therefore, the valid time steps are given by $G_t$, where $t \in [\frac{T}{2}, \frac{T}{2} + 1, \ldots, min(\frac{T}{2} + p_v T, T)]$, where $min(x,y)$ returns the minimum value between $x$ and $y$. Here, we used $p_v$ = 40\%.
	
	\item The probability function $\phi$ (Section~\ref{subsec:dif}) used is such that the chances of transmitting information to its neighbors is higher if the degree centrality of the transmitting node is higher than the receiving node. The probability $\phi_{ij}$ of node $i$ successfully transmitting information to node $j$ is given by:
	
	\begin{subequations}
		\begin{align}
			\mu &= \frac{d_{i}^{n}}{d_{j}^{n}};\\[7pt]
			\phi_{ij} &= m_c \psi \mathrm{e}^{\mu} \text{ , if j is a cyclic node;}\\
			\phi_{ij} &= \psi \mathrm{e}^{\mu} \text{ , otherwise;}
		\end{align}
	\end{subequations}
	
	where $d_{i}^{n}$ is the normalized degree of node $i$, $m_c$ is a constant for facilitating the transmission to cyclic nodes ($m_c > 1$), and $\psi$ is the diffusion rate. Here, we use $m_c = 3$ and $\psi = 0.002$;
	
	\item At time step zero, we select a set $P$ of nodes to start with the information (\textit{retained} nodes), as detailed in Section~\ref{subsec:dif}. We set $P = 0.02 N$, that is, we select 2\% of the nodes to start as \textit{retained} nodes. These nodes are selected as being the lowest degree nodes connected to at least one \textit{cyclic} node.
	
\end{itemize}

The training dataset is comprised of 500 artificial TVGs, while the test dataset is given by 50 artificial TVGs. Figure~\ref{subfig:train} shows how the degree varies in one of the artificial TVGs used in the training set. This figure shows how the degree of \textit{cyclic} nodes varies along with the time steps of the TVG, while normal nodes present only small fluctuations. All artificial TVGs present similar scatter plots (not identical).

Note that these artificial TVGs are built in order to present node behaviors that can be captured by the STIM network. After trained, the STIM is still able to capture these behaviors in other graphs, such as a real-world graph, as shown in Section~\ref{sec:real}.

\subsection{Training the STIM Agent}

STIM is trained according to the \textit{categorical algorithm}:
(i)~we first compute the $Q(s,a)$ for all nodes by running a forward pass over the model, which gives us matrix $Q_{all}$, from which we can obtain $Q(s,a)$ for all nodes by using Eq.~\ref{eq:dist_q}; (ii)~we then select the node (action) such that $a^* = arg max_a Q(s_{t+1}, a)$ or a random action with probability $e$, where $e$ is the exploration rate ($\epsilon$-greedy); (iii)~execute the selected action by propagating the data to the chosen node's neighborhood, which accepts the data with probability $\phi$; (iv)~obtain the reward; (v)~update the weights of the STIM network through backpropagation. The input vector $O_h$ in Figure~\ref{subfig:STIM} represents the one-hot encoding of the nodes chosen in the current batch. By multiplying $O_h$ with $Q_{all}$, we separate the distribution of the nodes selected as action. With these values, we compute the loss function $L_{ce}$ (Section~\ref{sec:rl}).
STIM is trained using an $\epsilon$-greedy policy, where we used an adaptive $\epsilon$ that varies between 0.8 to 0.3 along with the training procedure.

We use a decay factor for the learning rate, where its value is decayed after processing each batch. The decayed learning rate is given by:

\begin{equation}
	\alpha = \alpha \beta,
\end{equation}

\noindent where $\alpha$ is the learning rate and $\beta$ is the decay coefficient. We decay the learning rate until it reaches a minimum value $\alpha_{min}$. Refer to Table~\ref{tab:param} for the full list of parameters and their respective values used for the STIM network.

\begin{table}[ht]
	\centering
	\caption{Parameters used for training the STIM network. Refer to the text for a full explanation of each parameter.}
	\label{tab:param}
	\begin{tabular}{|l|c|c|}
		\hline
		\textbf{Parameter} & \textbf{Description}  & \textbf{\boldmath{Value}}	\\ \hline
		$B$             & batch size                    & 8             \\ \hline
		$\alpha $	    & learning rate                 & 0.0001       \\ \hline
		$\alpha_{min} $	& minimum learning rate         & 0.000001       \\ \hline
		$\epsilon$      & exploration rate              & 0.8 to 0.3    \\ \hline
		$\beta $	    & decay factor                  & 0.999       \\ \hline
		$L$	            & GE layers                     & 2       \\ \hline
		$\gamma$	    & discount factor               & 0.9       \\ \hline
		$C $	        & raw features                  & 4           \\ \hline
		$F $	        & embedding size                & 128         \\ \hline
		$V_{min} $	    & minimum sum of returns        & -1.0       \\ \hline
		$V_{max} $	    & maximum sum of returns        & 1.0       \\ \hline
		$n_a$   	    & number of atoms               & 11       \\ \hline
		$dif_{norm}$    & diffusion term weight         & 1.0       \\ \hline
		$inf_{norm}$    & influence term weight         & 0.5       \\ \hline
	\end{tabular}%
\end{table}

We also use experience replay, where we save to the replay buffer only successful episodes (percentage of $informed$ nodes $\geq 0.4$ after the end of the simulation). The experience replay buffer stores the last 25 positive episodes. We retrieve a random episode from this buffer with an interval of 3 simulations (run an episode from the buffer after processing 3 real simulations). This makes the learning process more stable.

\begin{figure}[!t]
	\centering
	\includegraphics[scale=0.55]{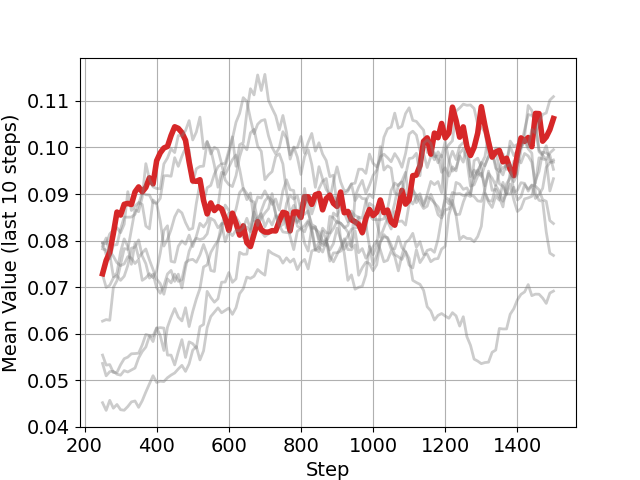}
	\caption{Training instances of the STIM agent, where each line represents a different instance. The best performing agent is depicted in red. The X-axis indicates the number of steps processed, where each step is given by a full simulation of a TVG. Therefore, the number of steps is the same as the number of TVGs processed (only the valid time steps of each TVG). The Y-axis indicates the percentage of $informed$ nodes after each simulation. Given the stochastic nature of the problem, we can observe great fluctuations in these values, so we plotted the moving average of the real values in order to smooth the curves.}
	\label{fig:train}
\end{figure}

Figure~\ref{fig:train} presents the training curves of ten instances of the STIM agent using the parameters of Table~\ref{tab:param}. Each curve represents a training instance. The red curve represents the best agent, which is the one used to generate the results presented in the following section.

\section{Performance Evaluation}
\label{sec:result}

In order to test the efficiency of the STIM network, we created two experiments: a diffusion process over a (i) set of artificial TVGs and (ii) a real-world TVG. We evaluate the performance of the model based on the percentage of $informed$ nodes after a given number of valid time-steps (also referred to as a simulation). We also report the mean percentage over several simulations, given that the simulation is highly stochastic and may report very different values for two identical simulations. 

In this section, we first present two agents used as a comparison baseline for the STIM agent. We then present the results of the STIM agent over the artificial TVGs, followed by the results over a real-world TVG. Finally, we perform a time complexity analysis of the STIM network.

\subsection{Greedy and Oracle Agents}
\label{sec:baseline}

We developed two additional agents in order to compare their results with the STIM agent: (i) a greedy agent used as a lower bound; and (ii)~a specialized agent built using specific knowledge of how the artificial TVGs are built, which is used as an upper bound. The former agent is called here the \textit{greedy} agent, while the latter is the \textit{oracle} agent. 

The \textit{greedy} agent uses a very simple strategy in order to select which of the \textit{retained} nodes it should activate at each time step: It selects the \textit{retained} node with the highest degree. This simple strategy serves as a lower bound agent, in the sense that no other agent must perform worse than the \textit{greedy} agent, given that it has a simple and low-cost solution to the maximum diffusion problem.

The \textit{oracle} agent is the opposite of the \textit{greedy} agent: It uses a highly specific strategy, tailored by using certain details of the creation of the artificial TVGs (Section~\ref{sec:artificial}). Therefore, this strategy represents an upper bound agent, since it uses privileged information regarding the creation process of the TVG used during training and testing. Therefore, it represents an impractical solution, given that in real-world scenarios, the details of how a TVG was created are typically unavailable. In general terms, the goal of the \textit{oracle} agent is to diffuse the information to a \textit{cyclic} node while it has a low degree, retain the data with this cyclic node until it reaches a high connectivity value, and only then selecting this node to diffuse the information. This strategy works for this set of artificial TVGs because diffusing an information to a high degree node is very difficult, given the probability function $\phi$ detailed in Section~\ref{sec:artificial}. Diffusing an information to a low degree node is easier. Therefore, we have to switch the state of a cyclic node from normal to retained while it has a low degree, and release this information only when it is well connected. The full strategy adopted by the \textit{oracle} agent is given by the following hierarchy of decisions (always execute the first top-most possible decision):

\begin{itemize}
	\item Select any type 4 node, if available;
	
	\item Select any type 2 node (if any), since these nodes are the ones connected to low connectivity cyclic nodes (type 1). Therefore, the nodes with a higher chance to transmit the information to a cyclic node are the type 2 nodes;
	
	\item Select the node with the highest degree.
\end{itemize}

\subsection{Experimental Results with Synthetic Networks}
\label{sec:test_net}


We tested the STIM, greedy, and oracle agents over the test dataset comprised of 50 artificial TVGs. For each agent, we executed 2000 simulations and saved the mean fraction of \textit{informed} nodes for all simulations. Note that the variation of informed nodes in each simulation suffers high fluctuations for all agents, given that due to the stochastic nature of the problem some simulations end up reaching only a few nodes, while other simulations may reach hubs, which in turn results in a higher information diffusion. Therefore, we only report the mean value for the 2000 simulations. We performed 5 instances of 2000 simulations and saved the mean value for each one, but in this case, the standard deviation observed for each instance of 2000 simulations is very small. 

Table~\ref{tab:results} presents the mean fraction of informed nodes of the 5~instances of the 2000 simulations over the test set of artificial TVGs for each of the tested agents. As we can see, the greedy model presents the worst results of the three agents, which is expected, given its limited strategy. The oracle agent has the best result, which is also expected, as stated in Section~\ref{sec:baseline}. The STIM agent then presents the intermediate result, being considerably better than the greedy agent, while reaching values comparable to the oracle agent. 

\begin{table}[ht]
	\centering
	\caption{Results obtained for the greedy, STIM, and oracle agents for the artificial TVG test set and the real-world TVG. The values for the test set correspond to the mean fraction of informed nodes per simulation for 5 sets of 2000 simulations (a mean of means). The values for the real-world set correspond to the mean fraction of informed nodes per simulation for 2000 simulations.}
	\label{tab:results}
	\begin{tabular}{|c|c|c|c|}
		\hline
		& \textbf{Greedy}  & \textbf{STIM}    & \textbf{Oracle}	\\ \hline
		\textbf{Test Set}       & 0.07182          & 0.11470             & 0.131615          \\ \hline
		\textbf{Real-World Set} & 0.03981          & 0.04997             & ---               \\ \hline
	\end{tabular}%
\end{table}

\begin{figure}[htp]
	\centering
	\includegraphics[scale=0.55]{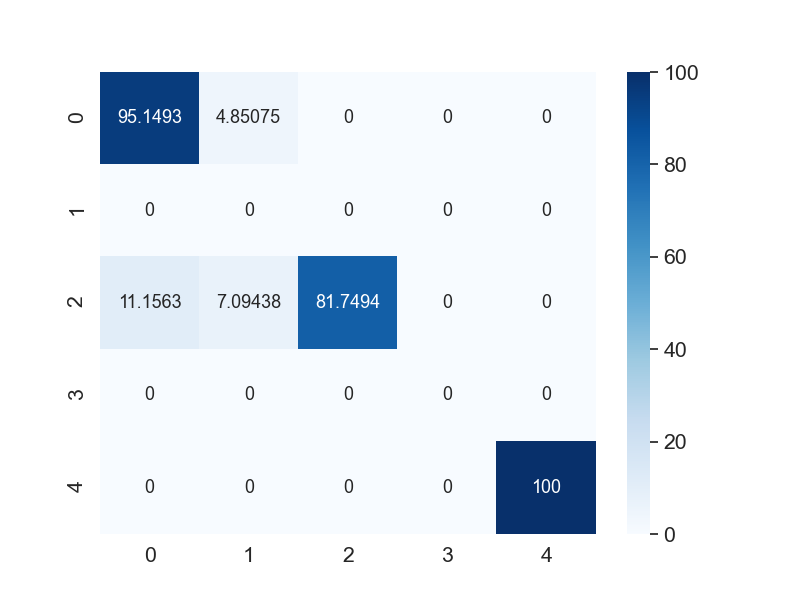}
	\caption{Confusion matrix representing how many times the STIM agent selected actions in accordance with the oracle agent. The value in cell $(i,j)$ corresponds to the percentage of times that the STIM agent selected a node of type $j$, while the oracle agent selected a node of type $i$ (considering both agents selected their actions at the same time and same state). For example, cell (2,1) corresponds to the number of times the oracle agent selected a node of type 2 while the STIM agent selected a node of type 1. The cells in the main diagonal represent the percentage of times both agents agreed in the selected node type.}
	\label{fig:conf_mat}
\end{figure}

To better analyze how the STIM agent learned its strategy, we tracked which node types it selected for each time step and compared these choices to the node types selected by the oracle agent at the same time step and same state. The results are presented in Figure~\ref{fig:conf_mat}, where we can see that both agents are usually in accordance with the node type that should be selected at each time step, shown by the high values in the main diagonal (see the figure description for more details). The main flaw of the STIM agent is that it sometimes does not select a node type 2 when it should, where it usually selects a node of type 0 in this case (shown by the relatively high value in cell (2,0)). By doing this, the chances of diffusing the information to a cyclic node decreases, which explains why the STIM agent did not perform as well as the oracle agent. This flaw is understandable, as the greatest challenge for efficiently diffusing the information in the artificial TVGs proposed is to correctly differentiate a node of type 0 from one of type 2. These node types are very similar, the only difference being that one has a cyclic node with a low degree (type 1 node) as a neighbor. Aside from this problem, the STIM agent managed to learn to retain the data when it reached type 1 nodes, only releasing it when their degree is high enough, i.e., after (almost) transforming to nodes of type~4. This can be observed in Figure~\ref{fig:conf_mat}, where we can see that the STIM agent rarely selects a type 1 node, shown by the low values in the second column.

\subsection{Experimental Results with Real-World Network}
\label{sec:real}


We now test our approach in a real-world TVG dataset.\footnote{http://snap.stanford.edu/data/email-Eu-core-temporal.html} The TVG used was built using email exchanged between a group of people over a span of 803 days.  The TVG has 986 nodes and 332334 temporal edges (across all time steps). We performed 2000 simulations over this network, using the diffusion parameters and rules like the ones used for the artificial TVGs. The only difference here is that we cannot separate each node in different classes, since there are no cyclic nodes. The starting nodes are selected as the top $P$ nodes with the lowest degree, where $P$ is defined similarly as in the artificial case. The degree variation of each node over the different time steps is presented in Figure~\ref{subfig:email}, where we can see a slightly different scenario than the one for the artificial TVG (Figure~\ref{subfig:train}).

Table~\ref{tab:results} presents the results for the e-mail TVG. We were not able to use the oracle agent for this scenario, given that this agent only works for the artificial TVGs, as previously detailed. As we can see, the STIM agent was still able to perform considerably better than the greedy agent, even though it was trained in a set of artificial TVGs with different dynamics. This shows how versatile the STIM framework is, where it is still able to generalize to new unseen scenarios.

\subsection{Time Complexity Analysis}
\label{subsec:time}

\begin{figure}[htp]
	\centering
	\includegraphics[scale=0.55]{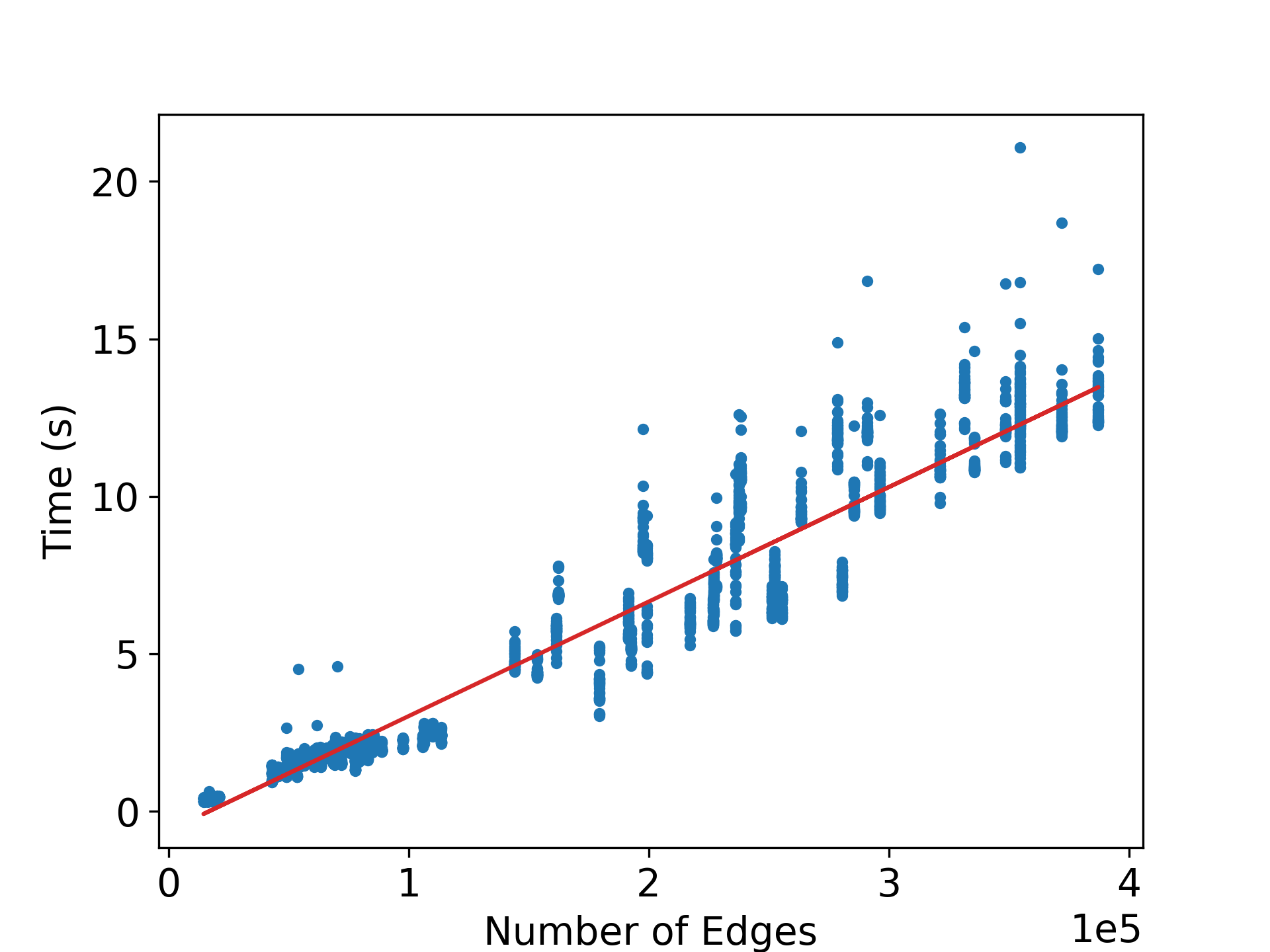}
	\caption{Time consumed for processing a batch of data retrieved from a specific time step of the TVG with different number of edges. Each data point represents a time measure performed for a specific time step of a TVG. The fitted line is shown in red, indicating a linear behavior.}
	\label{fig:time}
\end{figure}

We analyze the time complexity of the STIM network. We measured the time required by the proposal to process one batch of data for different sizes of TVGs. All experiments were performed on a machine with the following configurations: Ubuntu 18.04, Intel Core i7-7700K octa-core CPU with 4.20GHz, NVIDIA GTX 1080 with 8 GB of internal memory, 32 GB of RAM, using TensorFlow 1.14 with GPU support.

First of all, let us analyze the theoretical time complexity of the STIM network. As a starting point, it is important to note that we used sparse representations to deal with the larger matrices, such as the one-hot vector $O_h$ and the normalized adjacency matrix $\Tilde{D}^{-\frac{1}{2}} \Tilde{A} \Tilde{D}^{-\frac{1}{2}}$. STIM requires the sparse representation of the normalized adjacency matrix and the feature matrix as input, with time complexities of $O(|E|)$ (for sparse representation) and $O(N)$, respectively. The model then performs a series of matrix multiplications, where the upper bound is given by the matrix operations using the normalized adjacency matrix  $\Tilde{D}^{-\frac{1}{2}} \Tilde{A} \Tilde{D}^{-\frac{1}{2}}$, with dimensions $N \times N$, resulting in a time complexity of $O(N^2)$. However, given that we represent the $\Tilde{D}^{-\frac{1}{2}} \Tilde{A} \Tilde{D}^{-\frac{1}{2}}$ as a sparse matrix, this time complexity is reduced to the density of the matrix, i.e, $O(|E|)$. It is important to note that the STIM computes the $Q(s,a)$ distributions for all nodes in a single pass. Therefore, the time complexity of the STIM network is given by $O(N + |E| + c_p|E|) = O(|E|)$, where $c_p$ reflects the number of constant operations performed by STIM.  

Figure~\ref{fig:time} presents the time required to process one batch of data for TVGs of different sizes, where the TVGs used for this experiment varied from 1000 to 5000 nodes. Note that these results are for inference only, not training. Each data point represents a time measure performed for a specific time step of a TVG. For each point, we save the time required to process that particular time step, as well as the number of edges in that particular instant. We then fitted a line to the data, where we can see that, although there are some variations in the time measured (which is something common and expected), the line fitted and the original data present a linear behavior. Thus, we show that the STIM has a time complexity $O(|E|)$.

\section{Conclusions and Future Work}
\label{sec:conclusion}

We presented the STIM network, a model capable of predicting the best node to diffuse an information at a given time step in order to maximize the reach of such information in a TVG. The STIM model is trained using reinforcement learning and graph embedding techniques over a series of artificial TVGs built using non-trivial connectivity patterns. Our model was capable to learn such patterns and managed to efficiently propagate the information through the TVG over a set of limited time steps. It learned to retain the data within some specific nodes and release it only when their connectivity is high. It also learned to diffuse information from low degree nodes that are connected to cyclic nodes with low degree. Although this latter behavior was not fully learned, the STIM still managed to perform it very well. We compared the STIM with two other agents: A greedy agent that uses a simple strategy, and the oracle agent, that uses a highly specialized strategy built only for such artificial TVGs. The STIM managed to outperform the greedy agent, while achieving comparable results to the oracle agent, which is a great achievement, given that this latter agent uses an expert's knowledge of the TVGs, while the STIM learned it all from scratch. We then tested the STIM in a real-world graph with a different pattern than the ones used for training, and it still managed to perform well, which shows its generalization capabilities. STIM is also shown to have a time complexity of $O(|E|)$, indicating that the proposal may be used for larger graphs without becoming unfeasible.


The STIM represents a novel approach in the problem of spatio-temporal influence maximization in TVGs.
One can simply change some of the components of the STIM network, such as the graph embedding method, the RL method, or the reward function. Note that this makes the STIM very versatile, where one can adapt it to different scenarios by simply changing the reward function for example. This allows us to change the perspective of the problem and make it work for an epidemic or fake news spreading for example. Our method can also be adapted to consider budget constraints (as seen in previous approaches~\cite{Dyn_IC,node_sur}) by simply adapting the reward function such that it penalizes the agent for activating highly connected nodes, for example. These different perspectives constitute interesting future work that we plan to explore.

\section*{Acknowledgment}

This work has been partially supported by CAPES, CNPq, FAPERJ, and FAPESP. Authors also acknowledge the INCT in Data Science -- INCT-CiD.

\bibliographystyle{IEEEtran}
\bibliography{Ref}

\end{document}